\documentclass[conference]{IEEEtran}

\usepackage{cite}
\usepackage{amsmath,amssymb,amsfonts}
\usepackage{amsthm}
\usepackage{algpseudocode}
\usepackage{algorithm}
\usepackage{algorithmicx}
\usepackage{graphicx}
\usepackage[caption=false,font=footnotesize]{subfig}
\usepackage{textcomp}
\usepackage{xcolor}
\usepackage{makecell}
\usepackage{url} 
 
\def\BibTeX{{\rm B\kern-.05em{\sc i\kern-.025em b}\kern-.08em
    T\kern-.1667em\lower.7ex\hbox{E}\kern-.125emX}}

\begin{document}

\title{FedGA-Tree: Federated Decision Tree using Genetic Algorithm}

\author{\IEEEauthorblockN{Anh V. Nguyen}
\IEEEauthorblockA{
\textit{Northwestern University}\\
Evanston, USA \\
}
\and
\IEEEauthorblockN{Diego Klabjan}
\IEEEauthorblockA{
\textit{Northwestern University}\\
Evanston, USA \\
}
}
\maketitle

\begin{abstract}
In recent years, with rising concerns for data privacy, Federated Learning has gained prominence, as it enables collaborative training without the aggregation of raw data from participating clients. However, much of the current focus has been on parametric gradient-based models, while nonparametric counterparts such as decision tree are relatively understudied. Existing methods for adapting decision trees to Federated Learning generally combine a greedy tree-building algorithm with differential privacy to produce a global model for all clients. These methods are limited to classification trees and categorical data due to the constraints of differential privacy. In this paper, we explore an alternative approach that utilizes Genetic Algorithm to facilitate the construction of personalized decision trees and accommodate categorical and numerical data, thus allowing for both classification and regression trees. Comprehensive experiments demonstrate that our method surpasses decision trees trained solely on local data and a benchmark algorithm. 
\end{abstract}


\section{Introduction}
With rapid advancement of AI and machine learning, there are many concerns about data usage and privacy. Lawmakers worldwide have attempted to create incentives for companies to focus more on privacy in their model development, with key examples including the General Data Protection Regulations implemented by the European Union and the California Consumer Privacy Act.Federated Learning (FL) was introduced by Google as an approach for mobile devices to collaboratively solve a machine learning problem without sharing user's local data \cite{hard2018federated, kalloori2022cross}. In the FL framework, multiple clients contribute to solve a machine learning problem while maintaining their data locally. A global server helps aggregate information that clients deem fit to share, such as model weights, and construct an improved model. The two main scenarios of data distribution in FL are horizontal and vertical. In the former, clients have the same features but different set of samples while in the latter, clients have different features but the same set of samples. 

Currently, the main focus of the FL research community is on parametric, gradient-based models, yet there is an expanding body of literature that explores the use of decision tree models \cite{truex2019hybrid, wang2020scalable} \cite{du2002building, liu2020federated}. Decision tree (DT) is a powerful machine learning tool that is relatively easy to train and highly interpretable. Furthermore, its ensemble variants, such as XGBoost or Random Forest, have been shown to outperform neural networks on tabular data \cite{grinsztajn2022tree, gorishniy2021revisiting, shwartz2022tabular}. The most popular tree induction procedures are greedy algorithms, e.g. CART \cite{breiman2017classification}
and ID3 \cite{quinlan1986induction}, which rely on descriptive statistics of the data such as counts of classes to recursively build the tree top-down. Despite its advantages, the DT model can be susceptible to privacy breaches, as the decision nodes contain actual values of training data \cite{park2022evaluating, zhu2010understanding, zhu2010understanding}. In addition, by incorporating side information, an adversary can calculate the maximum posteriori estimate for a sensitive feature in the dataset \cite{park2022evaluating}.

Moderate efforts have been made to train tree-based models in the field of FL. Existing approaches often expand on greedy algorithms for tree induction and
incorporate differential privacy and encryption \cite{truex2019hybrid, wang2020scalable, du2002building, liu2020federated} to build a global decision tree for clients.  
Although these methods show promising results, there are certain shortcomings. Most prominently is the lack of personalization in the model. Furthermore, greedy algorithms themselves tend to be more vulnerable to local optima \cite{barros2011survey}, and when combined with differential privacy, their sequential top-down induction and iterative testing of all features can compound the errors from noisy estimates. This in turn constrains the possible maximum depth of the tree and restricts the application to only categorical data and classification problems.

To address these limitations, we propose a new approach to building personalized federated trees by leveraging Genetic Algorithm (GA) and using coarse aggregated information to ensure data privacy. GA is a metaheuristic method inspired by Darwinian evolution \cite{holland1992adaptation} that attempts to minimize an objective function by evolving an ensemble of candidate solutions. With a combination of exploitative and explorative operators, GA is equipped to perform a robust global search and avoid local optima \cite{barros2011survey}.
Specifically, our method utilizes the GA workflow to evolve a population of candidate solutions, which are binary tree structures with only decision node features and encoded as integer strings. Clients fit these structures to their local data to obtain personalized decision thresholds and leaf node labels and evaluate the fitted trees. The server then aggregates the fitness scores received from clients and applies genetic operators to produce a new generation of candidate trees. Our study has three main contributions:
\begin{itemize} 
    \item We are the first to propose a GA-based federated method that builds personalized binary decision trees in the horizontal FL framework. Our approach is compatible with categorical and numerical data, enabling the training of both classification and regression trees.
    \item Our method produces personalized trees, which is the first such approach based on DTs.
    \item We ensure the privacy of local data by removing feature thresholds and leaf labels from the trees, thus preventing attackers from inferring how the data are split. We also limit the exposure of direct descriptive statistics and only disclose coarse aggregated information (locally optimized maximum depths, coefficients of variation, and model evaluation scores) that provides a more abstract summary of the data.
    \item We validate our method with extensive numerical experiments and achieve superior performance compared to various approaches, including the state-of-the-art method by Truex et al. \cite{truex2019hybrid}. We also demonstrate the stability of our method in the case of partial client participation. 
\end{itemize}

The rest of the paper is organized as follows. We give an overview of decision trees and the related works in FL and outline the key components of GA in Section 2. In Section 3, we present the problem formulation and describe our tree-building algorithm with a discussion on privacy. Experiments and results are shown in Section 4, followed by a concluding discussion in Section 5.

\section{Related work}
In this section, we give an overview of decision tree models and discuss existing methods for training decision trees in the FL framework. We also present the general GA workflow and its key concepts. 

\subsection{Decision trees}
The decision tree (DT) is an important type of machine learning model that works well for medium-sized data and its advantages include simplicity, interpretability, and invariance with respect to feature scales. A decision tree comprises two types of nodes: decision nodes and leaf nodes. The former consists of a feature and threshold value and is used to route samples to a leaf node. The latter contains either the target class or a target value. 

The two most popular tree building algorithms for classification are CART \cite{breiman2017classification} and ID3 \cite{quinlan1986induction}. These greedy algorithms build a tree top-down by choosing the feature and associated value(s) that optimizes a chosen metrics, splitting the dataset, and continuing the procedure as needed. Leaf node labels are either the majority class (classification) or the average of the target values (regression). ID3 creates multi-way trees while CART selects a single threshold to create two branches in the tree. They use information gain or Gini impurity to decide a split. For regression, CART calculates mean squared errors ($MSE$s).


\subsection{Privacy-preserving DTs}
Because local data are incorporated into the model structure via decision nodes, DTs can pose significant privacy risks. Previous works address this issue by adding privacy-preserving protocols to existing greedy algorithms. These protocols often entail a combination of encryption (e.g. homomorphic encryption) and differential privacy (DP). In differentially private algorithms, the removal or addition of a single sample does not have a statistically significant impact on the output. In order to achieve differential privacy, random noise is added to the output, where the amount of noise is dictated by, and inversely proportional to, a predetermined privacy budget $\epsilon$. The lower the privacy budget, the more differentially private the algorithm is and the stronger the added noise needs to be. The Laplace Mechanism is a popular DP protocol.

To make DTs private, previous methods incorporate DP in the procedure by adding noise to class counts before calculating any metrics \cite{blum2005practical, friedman2010data}. Due to the composition theorem of DP \cite{dwork2009differential}, given a total privacy budget of $\epsilon$, the budget for each node at the same depth is $\epsilon/d$ where $d$ is the desired maximum depth of the tree. To determine the right split of a decision node, we need to further divide the privacy budget by the total number of potential splits, which scales not only with the number of features but also the number of unique values. As a result, these methods are either limited to categorical data and the classification task, or require discretization of continuous features and target values prior to training.

\subsection{Federated DTs}
Recent research on DTs in FL combines methods developed in the fields of privacy-preserving and distributed computing \cite{truex2019hybrid, wang2020scalable, du2002building, liu2020federated}. For the horizontal scenario, the state-of-the-art method is Truex et al. \cite{truex2019hybrid}, a protocol developed by IBM that combines differential privacy, encryption, and secure multiparty computation (SMC) to train a federated decision tree using the ID3 algorithm. The server queries clients for total counts and class counts to compute the information gain while clients add Laplace noise before returning a query output. A key contribution of this work is reducing the amount of noise added to the outputs by leveraging encryption and SMC. Specifically, Truex et al. consider a threat model in which the server and clients can collude among one another. Then given a minimum of $t$ honest clients, the required noise can be reduced by a factor of $t-1$.

Although Truex et al. \cite{truex2019hybrid} show promising results, there are certain shortcomings. First, despite reducing the required amount of noise, this new protocol is still limited to categorical data and classification only; regression problems or numerical data require discretization. Second, it is unclear whether the federated model consistently outperforms models that clients train locally without any federation. The authors only benchmark against the global tree without privacy guarantee and the traditional local DP protocol without SMC, and numerical experiments are applied to only one dataset. Third, although Truex et al. \cite{truex2019hybrid} experiment with varying the total number of clients in the FL system, the effect of partial client participation is not explored in the paper. During a round of communication, it is possible that not all clients are available to respond to the server's queries and engage in training. As a result, it is important to understand how the performance of a federated model changes in this scenario. Finally, the method by Truex et al. \cite{truex2019hybrid}, similar to all works in FL for DTs, builds a single global DT for all clients, yet this lack of flexibility can have a negative impact on the predictive power of the federated model. Our paper by employing GA addresses these limitations by enabling personalization and investigating the model performance with varied client participation and more datasets. 

There is also a rich literature on federated DTs that explores either the vertical scenario in data partitioning \cite{du2002building, liu2020federated} or extends FL to tree-based ensembles, i.e., bagging \cite{de2020dfedforest, giacomelli2019privacy, hou2021verifiable, kalloori2022cross, liu2020federated, yao2022efficient} and boosting \cite{chang2022fed, cheng2021secureboost, li2020practical, tian2020federboost, zhao2018inprivate}.  

\subsection{Genetic Algorithm}
\begin{figure}[!t]
    \centering
        \includegraphics[width=0.63\linewidth]{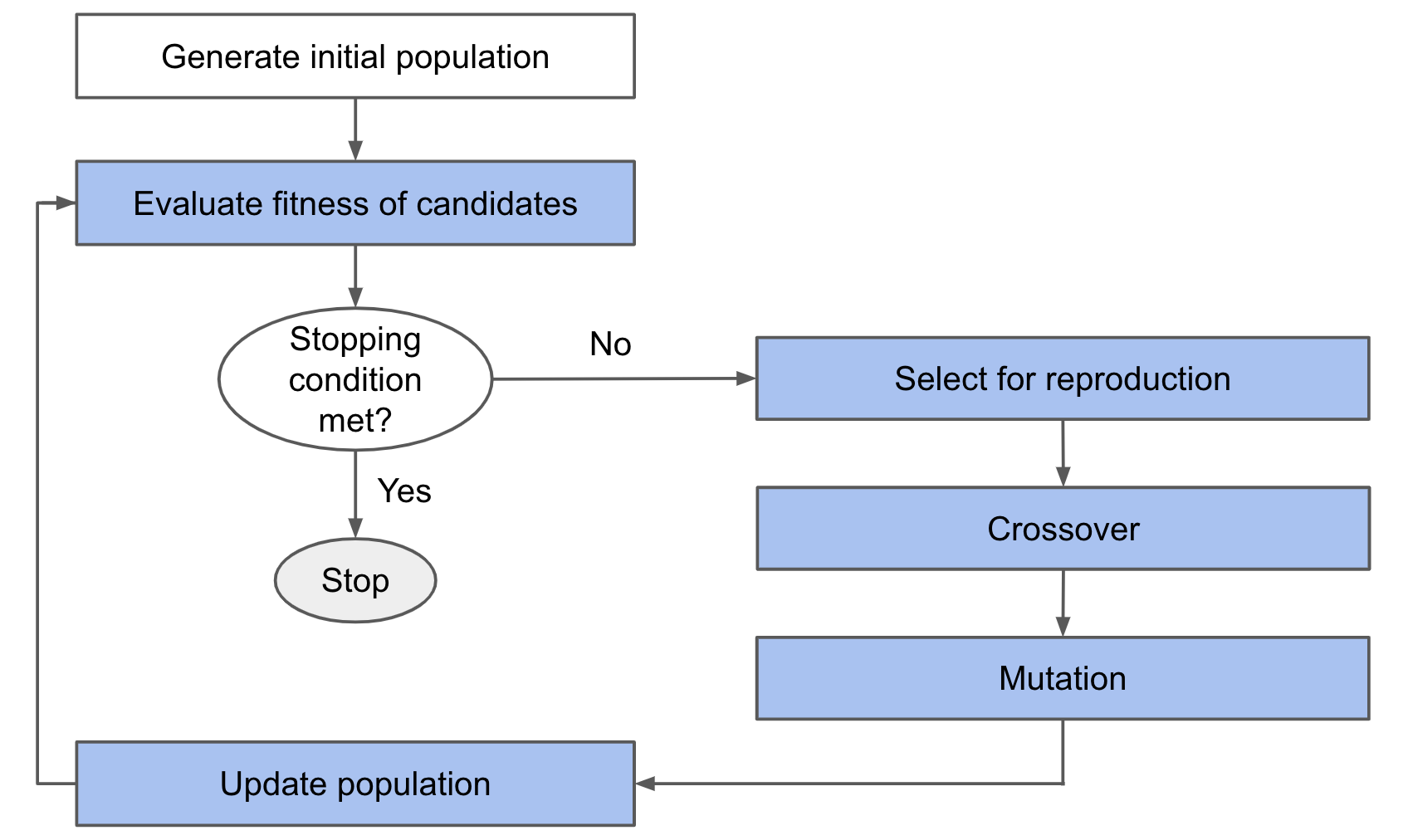}
        \caption{\footnotesize{Genetic algorithm workflow.}}
        \label{fig:ga}
\end{figure}

Genetic Algorithm (GA) is a highly flexible optimization method in which key procedures are customized according to the specific problem. The general workflow of GA is shown in Fig. \ref{fig:ga}. The algorithm starts with an initial population of candidate solutions. At each generation, GA evaluates the fitness of current candidates and selects only the top performers to generate new offspring. The basis of selection is the assumption that candidates with better fitness values contain better genetic matter to form the next generation. Tournament selection is one of the most popular schemes in GA \cite{barros2011survey}. A tournament of size $s$ means that $s$ individuals are chosen at random in a round robin fashion from the population and only the \textit{`fittest'} enters the pool for reproduction. To create new offspring, GA performs genetic operators, such as cross-over and mutation, on selected candidates. 

Barros et al. \cite{barros2011survey} provide a detailed review of the rich literature on the use of GA in decision tree induction. It should be noted that all of the existing works are limited to the centralized, nonfederated framework. In this paper, we adapt the GA workflow to the FL framework by introducing a new algorithm to effectively train federated DTs.  

\section{Method}
In this section, we present our federated algorithm, termed FedGA, which enables training of personalized binary decision trees in the horizontal FL framework. FedGA utilizes the GA workflow to evolve a population of structures, defined as search trees containing only decision features. By removing decision thresholds and leaf labels, we allow client personalization and ensure privacy by limiting the exposure of any descriptive statistics of local data. We differentiate between a structure which only has features in the corresponding search tree nodes and a decision tree which has features, thresholds, and leaf labels and can be used to make predictions. During each round of communication, clients get a population of candidate structures from the server, fit these structures to their local data to obtain personalized feature thresholds and leaf labels, and compute the fitness values (e.g. $F1$ score for classification and $MSE$ for regression). The server aggregates these fitness scores from clients and creates a new generation of structures accordingly. When training terminates, clients select from the final population of structures one that performs best on their local data. Thus, the final output of FedGA is a personalized decision tree for each client. Note that FedGA is compatible with any type of data and is applicable for both classification and regression.

The pseudocode is provided in Algorithm \ref{alg:fedga} in which steps 3-25 detail the initialization of FedGA and steps 26-43 lay out the training procedure. Finally, steps 44-49 give instructions on how clients can select their respective final decision tree. Fitness function $f$ is typically $F1$ or $MSE$ for regression.

FedGA starts with active clients performing $K$-fold cross-validation (CV) on local data to obtain the optimized maximum depth $d^i$ (step 6). In cross-validation training is performed and the validation metric is computed for each fold and each maximum tree depth in $[d_{min}, d_{max}]$. Clients also obfuscate the total number of samples with $\epsilon$ privacy by adding Laplace noise (step 7). The server determines the global maximum depth by taking the median of the proposed depths instead of the mean to avoid outliers and broadcasts $d$ to all clients (step 10). Clients then use CART to train a binary decision tree corresponding to the global maximum depth $d$ and send the structure in which its feature thresholds are removed to the server (step 12). The server initializes the GA population, incorporating the local structures received from clients (step 21). Details on the encoding of search tree structures and the $Initializer$ function to generate random structures are reported in Section \ref{subsect:encoding}. 

For regression, clients also send their votes on whether fitness scores of candidate structures should be calculated using the validation subset (train ratio $r < 1$) or the entire local training data (train ratio $r=1$) and whether to leverage elitism (steps 16-17). If more than half of the clients vote yes, the server adopts the respective strategy (steps 23-24).  

During each round of training, the server shares new candidate structures with the activate clients, who in turn fit the structures to their local data and compute the appropriate fitness scores (step 28). We discuss this $Evaluate$ procedure in more detail in Section \ref{subsect:evaluate}. In order to aggregate the fitness scores from clients, the server calculates the weighted average for each candidate search tree, using the noisy total sample counts (step 37). Training continues with the server updating the GA population by removing low-performing candidates (see Section \ref{subsect:update}) (step 39) and generating new offspring via the functions $Crossover$ and $Mutation$ (see Section \ref{subsect:operator}) (steps 40-41). Once the maximum number of generations is reached, each client receives the final set of candidate solutions from which to choose the top performer with respect to local data (steps 47-48).

Due to partial client participation, the server keeps track of which clients have participated in training before. As new clients join, they further send their perturbed sample sizes (steps 13, 32) as well as the local structures and corresponding fitness scores (step 31). The server adds these new structures to the GA population and continues the workflow.

We next discuss the details of our key GA operators in the following Sections \ref{subsect:encoding}, \ref{subsect:evaluate}, \ref{subsect:update}, and \ref{subsect:operator}, and the privacy aspect of FedGA in \ref{subsect:privacy}.  

\subsection{Tree structure encoding}\label{subsect:encoding}
We employ an encoding scheme that converts candidate structures into fixed-length integer strings. Each element in the string represents a decision node of a full binary search tree with maximum depth $d$ where the ordering follows a pre-order depth-first traversal of the tree. As a result, the length of the string is $2^d - 1$. We assume that each feature name and its metadata is hashed to an integer. These are search trees based on an implicit order defined by the algorithm, i.e. child nodes are ordered. Let $nil$ represent an integer that is not a hash value of any feature. The value of each element either corresponds to a decision node's feature hash value or takes the value $nil$ if a tree structure is incomplete such that the node is actually a leaf node or is nonexistent. 

\begin{figure}[!t]
    \centering
        \includegraphics[width=0.67\linewidth]{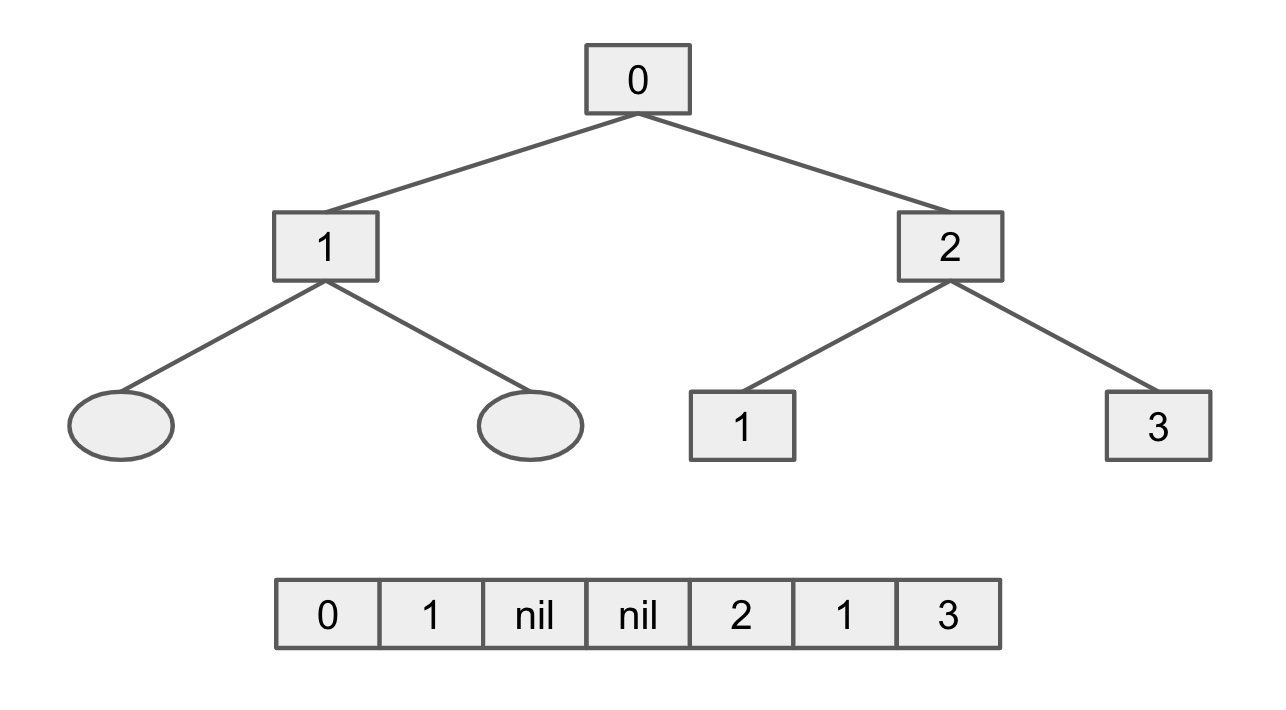}
        \caption{\footnotesize{Example of a search tree structure with a maximum depth of $d=3$ and the corresponding string representation.}}
        \label{fig:varencoding}
\end{figure}

Fig. \ref{fig:varencoding} illustrates an example of our encoding scheme for a structure with a maximum depth of $d=3$. The features are hashed to $0, ..., 3$ and $nil$ can be $-1$. Since the structure is incomplete, we need to add the appropriate paddings to ensure the correct ordering of the decision nodes. The advantages of converting structures into fixed-length strings include easy integration into the GA workflow and bloat control, that is, preventing trees to grow to excessively large depth, as in Genetic Programming \cite{luke2006comparison}. Guidotti et al. \cite{guidotti2024generative} use a similar representation, but instead of strings, they encode trees as matrices that contain feature thresholds. 

To decode, we can map the integer strings back into tree structures with a depth-first traversal. In the cases of detached subtrees, for example, if we replace $2$ by $nil$ in Fig. \ref{fig:varencoding}, we keep the connected tree that starts from the root and omit detached subtrees. 

To generate the initial GA population, the function $Initializer$ first creates random strings of integers by drawing from the set of features $\mathcal{F} = \{0, 1, ...,|\mathcal{F}|-1\}$ until the desired population size is reached. To encourage more diverse structures with varying distances from root to leaves, we randomly set a portion of the already selected decision nodes, as given by the leaf node ratio parameter $r_l$, to $nil$. 

\subsection{Evaluation of candidate structures}\label{subsect:evaluate}
After decoding the integer strings into search tree structures, clients have two main tasks: obtaining feature thresholds and calculating fitness scores. We follow the same procedure as in CART \cite{breiman2017classification} to select thresholds that minimize either Gini impurities (classification) or $MSE$ (regression). 

\begin{algorithm}[H]
\caption{Federated Genetic Algorithm (FedGA) Tree}\label{alg:fedga}
\begin{algorithmic}[1]

\State \textbf{Server Input:} GA population size $P$, number of GA generations $G$, privacy budget $\epsilon$, set of features $\mathcal{F}$, leaf node ratio $r_l$, tournament pressure $s$, number of node flip $m_1$, number of node swap $m_2$

\State \textbf{Client $i$ Input:} Dataset $\mathcal{D}^i$, $i\in[N]$; train ratio $r$, fitness function $f$, number of cross-validation folds $K$, range of possible tree depths $[d_{min}, d_{max}]$

\State \textbf{SERVER}
\State $\mathcal{P} \leftarrow \{\}$, $F \leftarrow \{\}$

\For{\textbf{CLIENT} $i$ in set of active clients $C_0$}

\State Sends $d^i \leftarrow CrossValidation(\mathcal{D}^i, K, d_{min}, d_{max})$


\State Sends sample size $n^i \leftarrow |\mathcal{D}^i| + Laplace(1/\epsilon)$

\EndFor

\State \textbf{SERVER}

\State Broadcasts $d \leftarrow Median(\{d^i, i \in C_0\})$ to all clients

\For{\textbf{CLIENT} $i$ in set of active clients $C$}


\State Constructs and sends local structure $T^i_d$

\State Sends sample size $n^i \leftarrow |\mathcal{D}^i| + Laplace(1/\epsilon)$ if $i \not\in C_0$

\If{\textbf{regression}}
\State $CrossValidation(\mathcal{D}^i, K, d_{min}, d_{max})$ if $i \notin C_0$

\State Sends vote on evaluation strategy $vote_1^i \in \{0, 1\}$ based on Eq. \ref{eq:evaluation}

\State Sends vote on elitism strategy $vote_2^i \in \{0, 1\}$ based on Eq. \ref{eq:elitism}
\EndIf

\EndFor
\State \textbf{SERVER}

\State $\tilde{\mathcal{P}} \leftarrow Initializer\big(\mathcal{F}, r_l, d, (P - |C|)^+\big) \cup \{T^i_d, i\in C\}$

\If{\textbf{regression}}
\State Broadcasts new train ratio $r=1$ to all clients if $\sum_{i\in C}vote^i_1 > |C|/2$

\State Uses elitism if $\sum_{i\in C}vote^i_2 > |C|/2$
\EndIf

\For{$g=1, ..., G$}

\For{\textbf{CLIENT} $i$ in set of active clients $\tilde{C}$}

\State $\tilde{F}^i=\{f^i_t, t\in \tilde{P}\} \leftarrow$ $Evaluate(\tilde{\mathcal{P}}, \mathcal{D}^i, r, f)$

\If{$i \not\in C$}
\State Constructs and sends local structure $T^i_d$

\State $\tilde{F}^i \leftarrow \tilde{F}^i \cup Evaluate(T^i_d, \mathcal{D}^i, r, f)$

\State Sends sample size $n^i \leftarrow |\mathcal{D}^i| + Laplace(1/\epsilon)$

\EndIf

\State Sends $\tilde{F}^i$ to the server
\EndFor

\State \textbf{SERVER}

\State $\tilde{F} \leftarrow \{f_t: f_t=\sum_{i \in \tilde{C}} f^i_t \cdot n^i/\sum_{j \in \tilde{C}} n^j, f^i_t \in \tilde{F}^i\}$

\State $\tilde{\mathcal{P}} \leftarrow \tilde{\mathcal{P}} \cup \{T^i_d, i \in \tilde{C} \setminus C\}$

\State $\mathcal{P}, F \leftarrow Update(\mathcal{P}, \tilde{\mathcal{P}}, F, \tilde{F})$

\State $C \leftarrow C \cup \tilde{C}$

\State $\tilde{\mathcal{P}} \leftarrow Crossover(\mathcal{P}, F, s)$

\State $\tilde{\mathcal{P}} \leftarrow Mutation(\tilde{\mathcal{P}}, m_1, m_2, \mathcal{F})$
\EndFor

\State\textbf{SERVER}

\State Broadcasts $\mathcal{P}$ to all clients;

\For{\textbf{CLIENT $i=1,...,N$ in parallel}}

\State $F^i \leftarrow Evaluate(\mathcal{P}, \mathcal{D}^i, r, f)$

\State Selects best performing decision tree
\EndFor

\end{algorithmic}
\end{algorithm}

The following applies only to regression and during the initialization phase (steps 3-25). The different treatment from classification is needed, as building a regression tree is typically more difficult than a classification tree. In fact, the regression tree model predicts a specific value and uses metrics that are not bounded, such as MSE. Thus, small deviations can be heavily penalized, resulting in regression trees having more sensitivity to noise and instability with respect to evaluation scores, compared to classification trees. 
Therefore, to calculate fitness scores, we use either the entire local training set or the validation subset, depending on the characteristics of the dataset. With the former strategy, clients use all of the local data for both tasks. With the latter strategy, clients create a train/validation split and use the training subset to obtain full decision trees from structures and the validation subset to compute the fitness scores. 

In determining which strategy to employ, we characterize a dataset by how noisy the data are and how complex the interactions among features are. Recall that we require clients to perform a $K$-fold CV on the local data to select the maximum depth. If the data are noisy, the CV scores can vary significantly across different folds. Furthermore, deeper trees tend to better explain the data with complex interactions among features. 

Suppose that for each client $i$, the cross-validation score corresponding to a maximum depth $j$ and fold $k$ is denoted by $v^i_{kj}$. Let $\phi^i_j$ be the coefficient of variation of values $\{v_{kj}^i, k \in [K]\}$ and $\theta^i_k$ be the coefficient of variation of values $\{v^i_{kj}, j\in [d_{min}, d_{max}]\}$. These values are computed in steps 6 and 15 for each client $i \in C \cup C_0$. We characterize the amount of noise in a dataset by the coefficient of variation, $\phi^i_d$, of the validation scores across different folds at $j=d$ where $d$ is the global maximum depth chosen by the server. In addition, we define the complexity of feature interactions in a dataset as the average coefficient of variation of the validation scores across different maximum depths, i.e. the mean of $\{\theta^i_k, k \in [K]\}$. The choice of these metrics is due to the fact that coefficient of variation is a dimensionless measure, thus well-suited for comparison between datasets with varying units. We use the $R^2$ score as the cross-validation metric.  

After calculating these two metrics, clients determine which strategy should be used. We find that the first strategy of using all of the local data is more beneficial for datasets with less noise ($\phi^i_d < \phi^*$) and more complex interactions between features ($\theta^i > \theta^*$) where $\phi^*$ and $\theta^*$ are thresholds obtained experimentally. To this end, we set
\begin{equation}
vote_1^i =
\begin{cases}
1 & \text{if } \phi_d^i < \phi^* \text{ and } \theta^i > \theta^* \\
0 & \text{otherwise.}
\end{cases}
\label{eq:evaluation}
\end{equation}
The server selects the strategy with more than half of the votes and relays that decision to the clients.

\subsection{Updating the GA population}\label{subsect:update}
Before advancing to the next generation, GA combines the parent population and its offspring pool to eliminate low performers ranked by the fitness scores. 

For regression, if the target variable has high cardinality, that is, a high percentage of unique values, the server incorporates elitism, keeping the set of tree structures received from clients in every generation. Otherwise, we employ the traditional update strategy. 

Let $\mu^i$ denote the percentage of unique values for the target variable in the local data of the client $i$. We set 
\begin{equation}
vote_2^i = 
\begin{cases}
1 & \text{if } \mu^i > \mu^* \\
0 & \text{otherwise}
\end{cases}
\label{eq:elitism}
\end{equation}
where $\mu^*$ is determined experimentally. If more than half of the active clients vote for elitism, the server implements the strategy in each GA iteration. 

\subsection{Crossover and mutation}\label{subsect:operator}
In FedGA, the implementation of crossover is straightforward due to the string representation. We employ the tournament selection strategy with pressure to select the reproduction pool and perform the popular one-point crossover. On the other hand, mutation requires more careful design to ensure effective exploration, as simply changing a feature may not be enough to escape the local search space. We devise three mutation strategies - node flip, node swap, and substring swap - and randomly apply one in each generation of GA. 

Node flip is a multi-point mutation in which we randomly select a number of positions in the integer string, given by the parameter $m_1$, and replace the current values with either a different feature or $-1$. Node swap involves the exchange of values between two random positions, and this procedure can be repeated $m_2$ times for better exploration. Finally, for substring swap, we split the candidate into three substrings and randomly reorder them. In implementation, we set $m_1$ and $m_2$ as some percentage of the string length.  

\subsection{Privacy discussion}\label{subsect:privacy}
We consider an honest-but-curious scenario, similar to Truex et al. \cite{truex2019hybrid}, in which all parties follow the instructions but still attempt to infer information related to clients' data. Unlike \cite{truex2019hybrid}, we do not rely on DP to explicitly guarantee privacy, except in the one-time communication of clients' sample size, because FedGA is not built on direct descriptive statistics of local data, but instead uses coarse aggregated information. 

Specifically, the only descriptive information about client data that the server has access to is the (perturbed) total sample size, the list of features, the locally optimized maximum depth, and the coefficients of variation related quantities ($\phi^i_d$ and $\theta^i$). The sample size is a direct statistics and therefore we implement DP to ensure privacy for clients. Furthermore, it is common to assume that the list of features is global knowledge in the horizontal FL setting. The maximum depth and the coefficients of variation limit a potential membership inference attack because they are coarse metrics aggregated over multiple folds and provide an abstract description of the data. 

During training, our procedure does not involve any additional descriptive statistics of local data and only requires the disclosure of evaluation scores, $F1$ or $MSE$. We want to emphasize that the evaluation is done by clients on local data, and the server does not query the evaluation of search tree structures for additional data samples. As a result, the server can observe how a change in decision node features can affect the evaluation and, for example, infer that one feature has more impact than others. However, there is no knowledge about the specific value that the feature can take on and how the data are split. 

\section{Experiments and results}
In this section, we demonstrate the performance of FedGA in both classification and regression as well as with full and partial client participation. 
We benchmark against the centralized tree, which is trained using centrally aggregated data, and local trees, which are trained locally by clients. For classification, we further benchmark against the DP-based tree building algorithm by Truex et al. \cite{truex2019hybrid}, referred to as IBM tree henceforth. Because the method is originally developed for multiway trees with the ID3 algorithm, we adjust the IBM protocol to build binary trees with CART instead to ensure fair juxtaposition with FedGA. To capture the effect of randomization in both methods, we execute 10 independent runs given a train/test split and report the averages across these runs and clients. 

\subsection{Experimental settings}
\subsubsection{Data}
We conduct experiments on 14 classification datasets and 14 regression datasets (Table \ref{table:datasets}). The first two classification datasets, \texttt{nusery} and \texttt{adult}, are provided in the IBM Federated Learning repository\footnote{\url{https://github.com/IBM/federated-learning-lib/tree/main/examples/id3_dt}} that hosts the method developed by Truex et al. \cite{truex2019hybrid}. The remaining datasets are curated from Grinsztajn\footnote{\url{https://huggingface.co/datasets/inria-soda/tabular-benchmark}} et al. \cite{grinsztajn2022tree}, which explores how tree-based models can still outperform deep neural networks on tabular data, but are limited to those with fewer than $75000$ samples and fewer than $30$ features. Since IBM trees are constructed with DP and are highly sensitive to the cardinality of features, we categorize any numerical features in classification datasets using the \texttt{KBinsDiscretizer} function from the \texttt{scikit-learn} package. Details on the discretization are deferred to Appendix \ref{appendix:preprocessing}. 

Each dataset is randomly divided into train/test sets with proportion $90\%/10\%$ and the latter is common among clients for comparison. We partition the training data into disjoint subsets by randomly assigning samples to a client. This results in clients having approximately equal number of samples and roughly the same distribution of classes (classification) or target values (regression). To account for the noise in the data and guarantee the robustness of our results, we generate 10 independent train/test splits and repeat the same experiments $10$ times for each split. In each iteration of the algorithm, we generate active clients uniformly at random where $c$ denotes the percentage of active clients.

\begin{table}[!t] 
\caption{Datasets. The table contains information on number of samples $n$ and number of features $|\mathcal{F}|$.}
\label{table:datasets}
\centering
\begin{tabular}{c r r | c r r}
\hline
\hline
\multicolumn{3}{c}{Classification} & \multicolumn{3}{c}{Regression} \\
\hline
Dataset & $n$ & $|\mathcal{F}|$ & Dataset & $n$ & $|\mathcal{F}|$ \\
\hline
\hline
\texttt{nursery} & 12960 & 8 & \texttt{analcatdata} & 4052 & 7 \\
\hline
\texttt{adult} & 48842 & 14 & \texttt{abalone} & 4177 & 8 \\
\hline
\texttt{california} & 20634 & 8 & \texttt{bikesharing} & 17379 & 11 \\
\hline
\texttt{compas} & 4966 & 11 & \texttt{brazil} & 10692 & 11\\
\hline
\texttt{creditcard} & 13272 & 21 & \texttt{cpu} & 8192 & 21\\ 
\hline
\texttt{diabetes} & 71090 & 7 & \texttt{elevators} & 16599 & 16\\
\hline
\texttt{electric} & 38474 & 8 & \texttt{house16H} & 22784 & 16\\
\hline
\texttt{eyemvt} & 7608 & 23 & \texttt{houses} & 20640 & 8\\
\hline
\texttt{heloc} & 10000 & 22 & \texttt{housesales} & 21613 & 17\\
\hline
\texttt{house16H} & 13488 & 16 & \texttt{miami} & 13932 & 13\\
\hline
\texttt{kaggle} & 16714 & 10 & \texttt{soil} & 8641 & 4\\
\hline
\texttt{marketing} & 10578 & 7 & \texttt{sulfur} & 10081 & 6\\
\hline
\texttt{pol} & 10082 & 26 & \texttt{wine} & 6497 & 11\\
\hline
\texttt{telescope} & 13376 & 10 & \texttt{yprop} & 8885 & 42\\
\hline
\hline
\end{tabular}
\end{table}

\subsubsection{Benchmarks}
To demonstrate the efficacy of our method, we use the following benchmarks:
\begin{itemize}
    \item Centralized tree, trained by a central server that has access to all client data;
    \item Local tree, trained by each client locally and in silo, i.e., without any collaboration with other clients; and
    \item IBM tree, trained with the DP protocol in \cite{truex2019hybrid} and adjusted for binary trees. This benchmark is only applied to classification datasets since the method is limited to categorical data only. 
\end{itemize}

Note that the IBM method is developed for multiway trees, while FedGA is specific to binary trees. We replace the ID3 procedure in \cite{truex2019hybrid} with CART so that the IBM method produces a single binary tree instead. We fix the privacy budget at $\epsilon=1$.
As shown in Truex et al. \cite{truex2019hybrid}, the performance of IBM trees for $\epsilon \geq 1$ is relatively stable, while smaller values of
$\epsilon$ exhibit a noticeable degradation in results. Thus, we pick $\epsilon=1$ to balance between performance and privacy for the IBM method. We also assume that all clients are honest and non-colluding. To determine the maximum depth for IBM trees, we randomly pick three datasets (\texttt{hou16H}, \texttt{kaggle}, and \texttt{marketing}) and fine-tune the depth within the range $[2, 15]$. We find that a maximum depth of $6$ yields the best performance on average and fix this depth for all datasets. We include the results for training IBM trees with the median depth strategy of FedGA and the half strategy of the original paper where the maximum depth is chosen as $|\mathcal{F}|/2$, in Appendix \ref{appendix:ibmdepth}.

With the centralized and local trees, we perform $K$-fold cross-validation ($K=5$) on the aggregated and local data respectively, to select the maximum depth. Note that the local trees can have the same structure as those used to initialize GA if the locally optimized depth is the same as the depth selected by FedGA. We refer to the latter as local structures to differentiate them from the local decision trees.

Regarding metrics, we calculate the $F1$ score on the test data for classification and $MSE$ for regression. Since maximum depths can vary across different methods, we quantify the complexity of a tree by the number of decision nodes. Moreover, because we incorporate local structures in the initialization, we evaluate whether FedGA is able to navigate the search space effectively or it is stuck with the initial local structures and the search thus becomes futile. Hence, we calculate the normalized tree edit distance between the list of hashed decision features of a client's final FedGA tree and that of the respective local structure. 

Finally, to better understand the client-level benefit of FL, we compute the following metric for each client
\begin{equation}
    \Delta^i = \Big(\dfrac{\text{Test score of federated tree}}{\text{Test score of local tree}} - 1\Big) * 100,
\end{equation}
and obtain $\Delta = Mean(\{\Delta^i, i\in[N]\})$ where $N$ is the total number of clients in the FL system. This metric measures the average percentage increase in the test score of federated trees compared to local trees. For classification, the desired $\Delta$ in $F1$ score is greater than $0$, while for regression, negative values are preferred for $MSE$.

\subsubsection{Implementation}
We limit the range of possible tree depths $[d_{min}, d_{max}]$ to $[2, 15]$ to maintain interpretability. We report the remaining  hyperparameters of FedGA in Table \ref{table:hyperparameters}, which are obtained from manual fine-tuning. 

\begin{table}[!t] 
\renewcommand{\arraystretch}{1.1}
\caption{Hyperparameters in FedGA.}
\label{table:hyperparameters}
\centering
\begin{tabular}{c c c c}
\hline
\hline
$P$ & $G$ & $\epsilon$ & $r$  \\
\hline
$20$ (classification) or $40$ (regression) & $100$ & $1$ & $0.8$\\
\hline
\hline
\end{tabular}
\begin{tabular}{c c c c c c c}
$r_l$ & $s$ & $m_1$ & $m_2$ & $\phi^*$ & $\theta^*$ & $\mu^*$\\
\hline
$0.01$ & $3$ & $0.05$ & $0.05$ & $0.25$ & $0.02$ & $80$\\
\hline
\hline
\end{tabular}
\end{table}

All codes are written in Python version 3.11.0 and experiments are performed on a heterogeneous high-performance computing cluster consisting of Intel Cascade Lake $6230$, Ice Lake $6338$, and Emerald Rapids $8592+$
nodes.

\subsection{Results}
\subsubsection{Classification}
\begin{figure}[!t]
\centering
\subfloat[$\Delta$ in $F1$ score of federated trees compared to local trees.]{
\includegraphics[width=\linewidth]{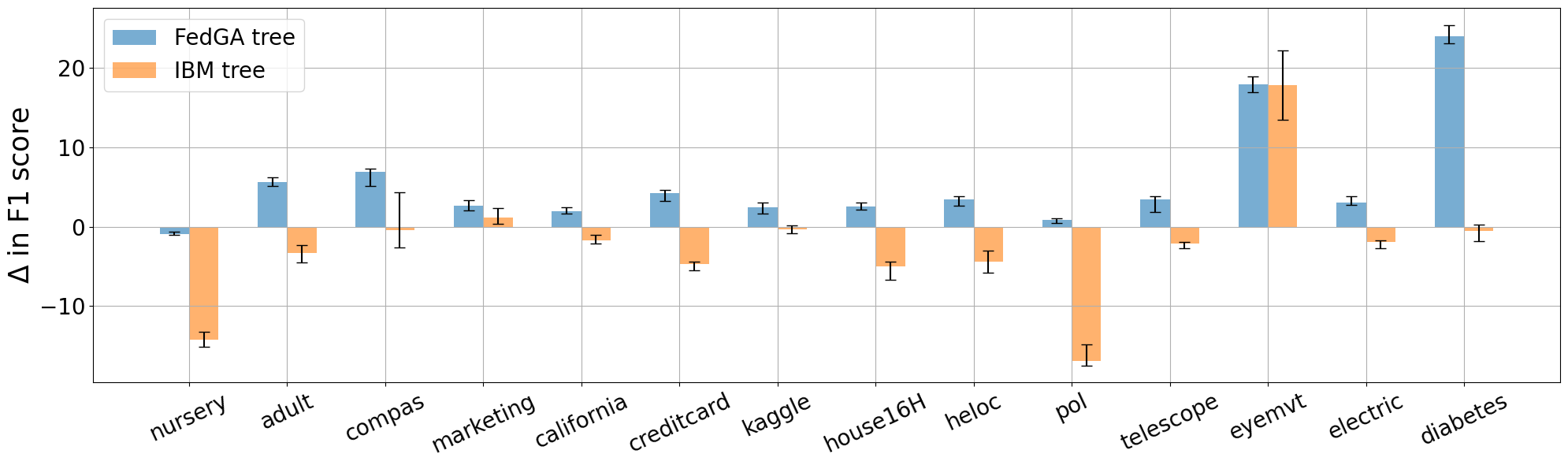}}
\hfil
\subfloat[Percentage of clients for which the federated trees outperform local trees.]{\includegraphics[width=\linewidth]{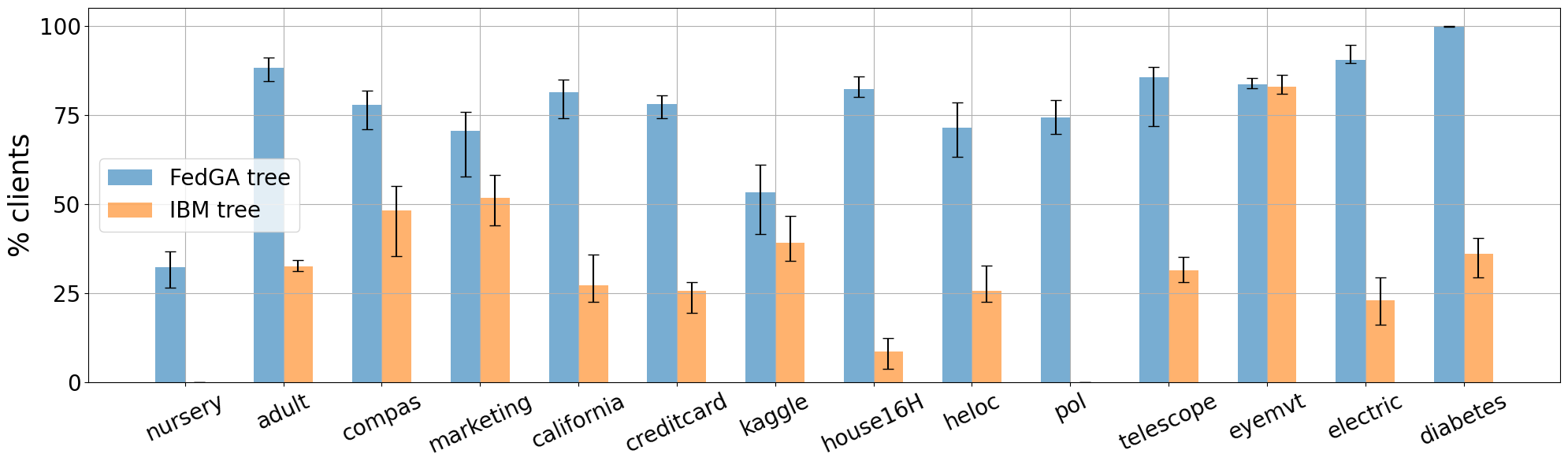}}
\caption{\footnotesize{Performance of federated trees compared to local trees for different datasets across 10 random train/test splits. (a) The medians and interquartile ranges (IQRs) of $\Delta$ in $F1$ score. (b) The median percentages of clients for which federated trees perform better as well as the corresponding IQRs. The $\Delta$'s and percentages of clients are averaged across independent runs for each train/test split.}}
\label{fig:classification_iid_full}
\end{figure}

An important aspect in evaluating a federated model is how much the results are improved relative to locally trained models. We illustrate this comparison in a 20-client FL system with $c=100\%$ for different datasets in Fig. \ref{fig:classification_iid_full} with two metrics, the $\Delta$ in $F1$ score and the percentage of clients for which federated trees perform better. We later vary $N$ and $c$. These two values of $N$ and $c$ are selected so that each client has at least $200$ samples to ensure meaningful local training. The experimental results in Fig. \ref{fig:classification_iid_full}(a) offer compelling evidence for the superior performance of our method, compared to IBM and local trees. Furthermore, for every dataset except \texttt{nursery}, FedGA has more than $50\%$ of clients benefiting from federated training. Across all datasets, FedGA achieves a $\Delta=5.58$ on average, while IBM trees result in a $\Delta=-2.58$. Similarly when averaged across datasets, $75\%$ of clients show better performance compared to local models with FedGA trees but only $31\%$ do with IBM trees. 

\begin{figure}[!t]
\centering
\subfloat[Number of decision nodes.]{\includegraphics[width=\linewidth]{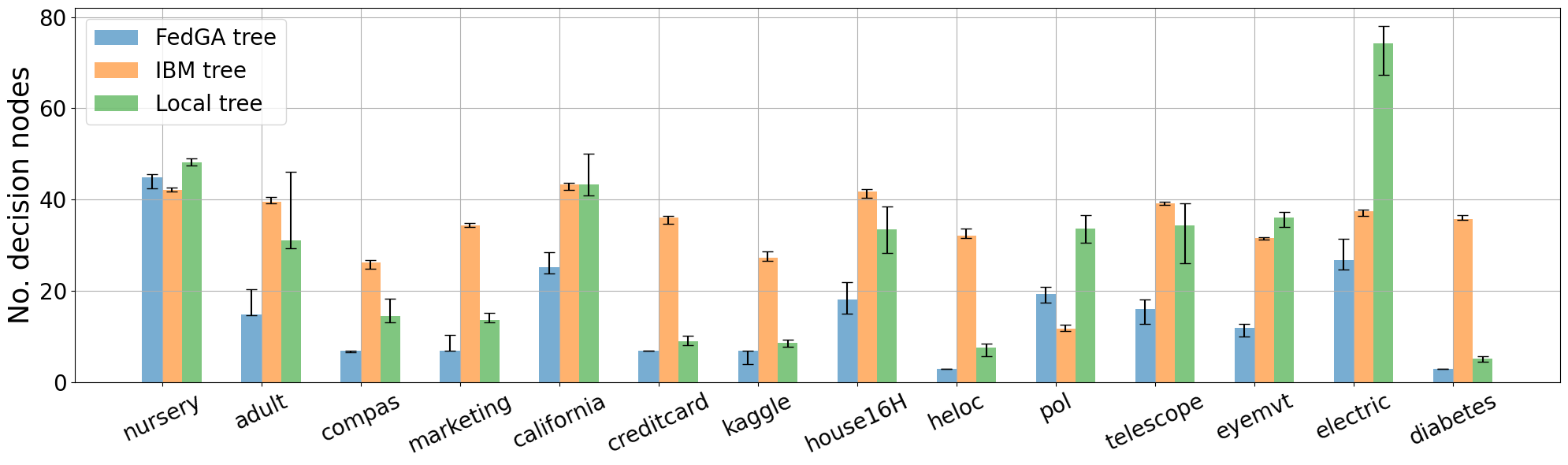}}
\hfil
\subfloat[Normalized tree edit distance between clients' FedGA trees and their corresponding local trees.]{\includegraphics[width=\linewidth]{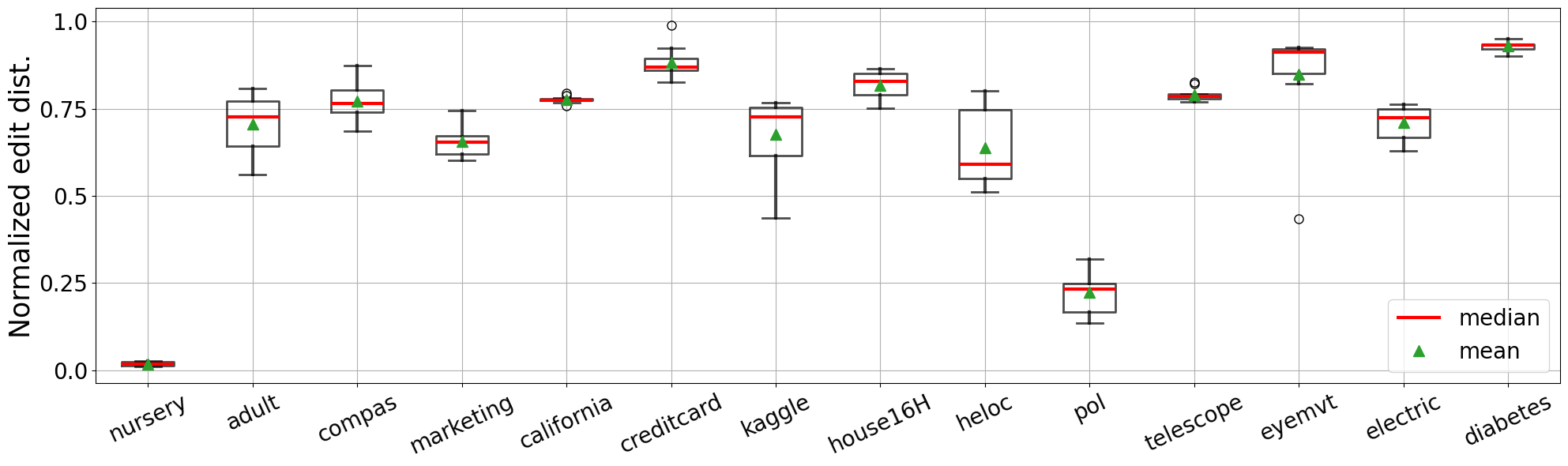}}

\caption{\footnotesize{Comparison of tree structures for different datasets across 10 random train/test splits. (a) The medians and IQR's of the number of decision nodes averaged across clients and independent runs of the FedGA and IBM method. (b) Boxplots of the normalized tree edit distance between clients' FedGA trees and their corresponding local structures, averaged across independent runs.}}
\label{fig:classification_iid_full_tree}
\end{figure}

We also present an analysis of the complexity of FedGA trees in Fig. \ref{fig:classification_iid_full_tree}. Evidently, the trees produced by FedGA are less complex than the local trees, which likely leads to better generalizability and explains the increase in the $F1$ score. In contrast, IBM trees consist of considerably more decision nodes than local trees, explaining their modest performance. In Fig. \ref{fig:classification_iid_full_tree}(b), we show the normalized tree edit distance between the FedGA outputs and the local structures in the initial GA population. On average, FedGA trees are significantly different from their corresponding local structures as normalized edit distances are greater than $50\%$ for most datasets ($12/14$). This indicates that FedGA is able to navigate the search space effectively and is not easily trapped in local optima. 

\begin{figure}[!t]
\centering
\subfloat[Test scores for one train/test split.]{\includegraphics[width=0.65\linewidth]{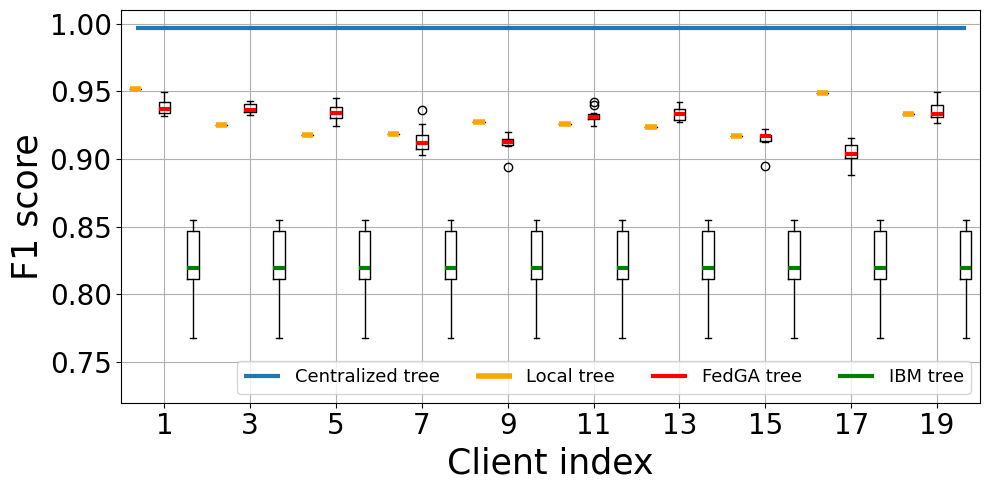}}
\hfil
\subfloat[Test scores averaged across clients for 10 random train/test splits.]{\includegraphics[width=0.65\linewidth]{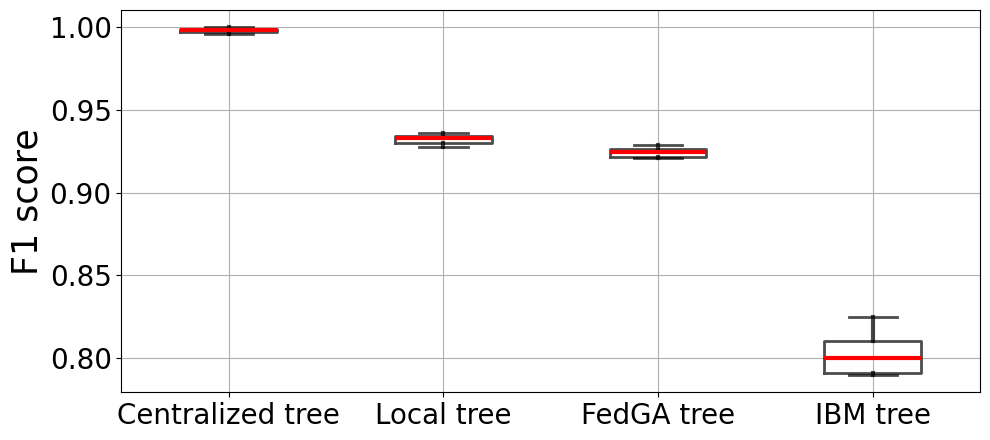}}
\caption{\footnotesize{Performance of different methods on the \texttt{nursery} dataset. (a) The IQRs of test scores for one random train/test split of the data. Every other client is displayed. (b) The distribution of test scores for different train/test splits of the data. The test scores of FedGA and IBM trees are averaged across clients and independent runs.}}
\label{fig:nursery}
\end{figure}

We give further insight into the performance of FedGA for the \texttt{nursery} dataset, which is the only one yielding negative $\Delta$ for both FedGA and IBM trees. In Fig. \ref{fig:nursery}(a), it is evident that the IBM tree induction procedure is more sensitive to randomization, compared to FedGA. We point out that IBM generates the same tree for each client in each run. Furthermore, as shown in Fig. \ref{fig:nursery}(b), the IBM trees clearly underperform compared to the local trees, whereas the performance gap for FedGA is much less significant. 

\begin{figure}[!t]
\centering
\subfloat[FedGA trees]{\includegraphics[width=\linewidth]{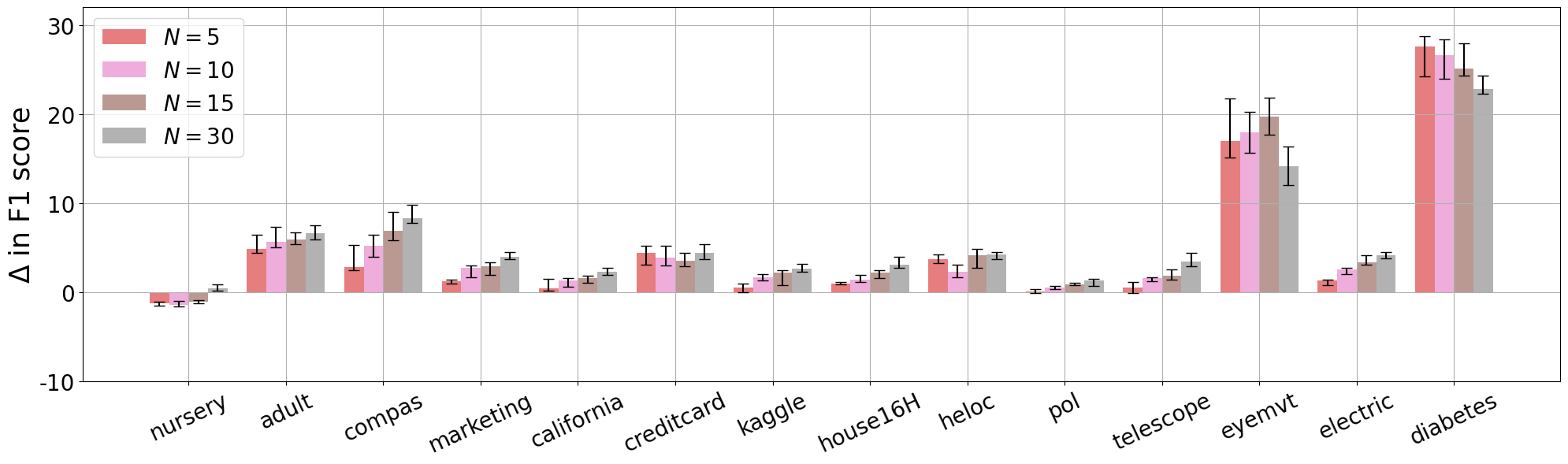}}
\hfil
\subfloat[IBM trees]{\includegraphics[width=\linewidth]{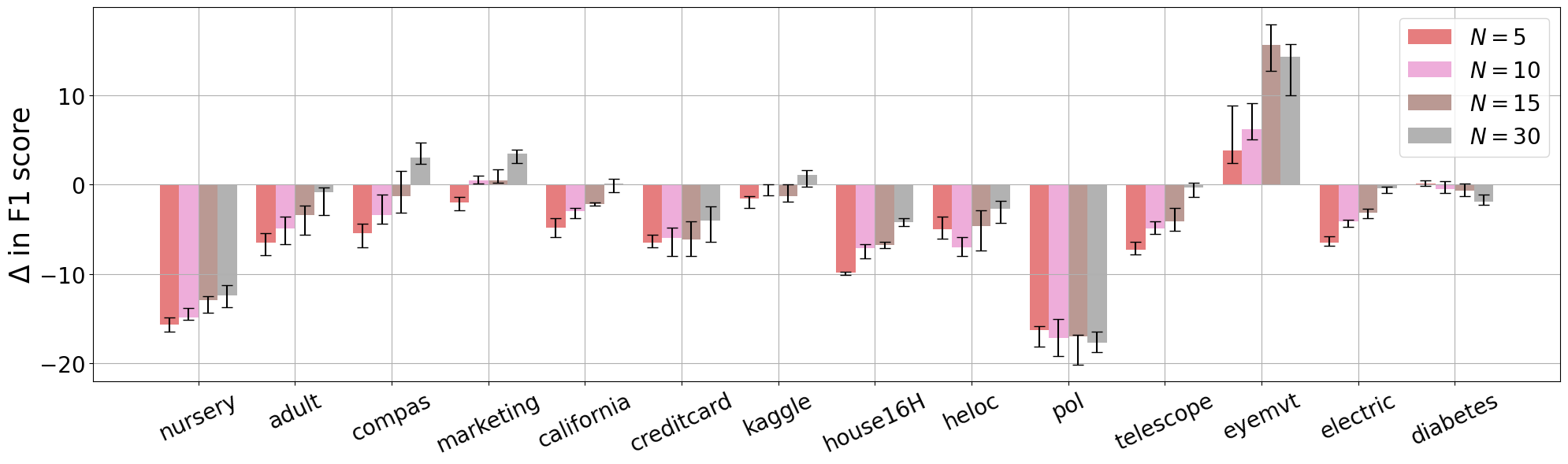}}

\caption{\footnotesize{$\Delta$ in $F1$ score across random train/test splits for FedGA and IBM trees as the total number of clients in the system varies.}}
\label{fig:classification_iid_vary_clients}
\end{figure}
Next, in Fig. \ref{fig:classification_iid_vary_clients} we present the results of FedGA as the total number of clients varies $N \in \{5, 10, 15, 30\}$. The total number of clients in a system dictates the size of the entire FL system, and as the number of clients increases, the size of local data decreases. Fig. \ref{fig:classification_iid_vary_clients} illustrates the $\Delta$ in $F1$ score with full client participation. We observe a similar pattern as $N=20$, i.e. FedGA consistently outperforms IBM trees.

\begin{figure}[!t]
\centering
\subfloat[FedGA trees]{\includegraphics[width=\linewidth]{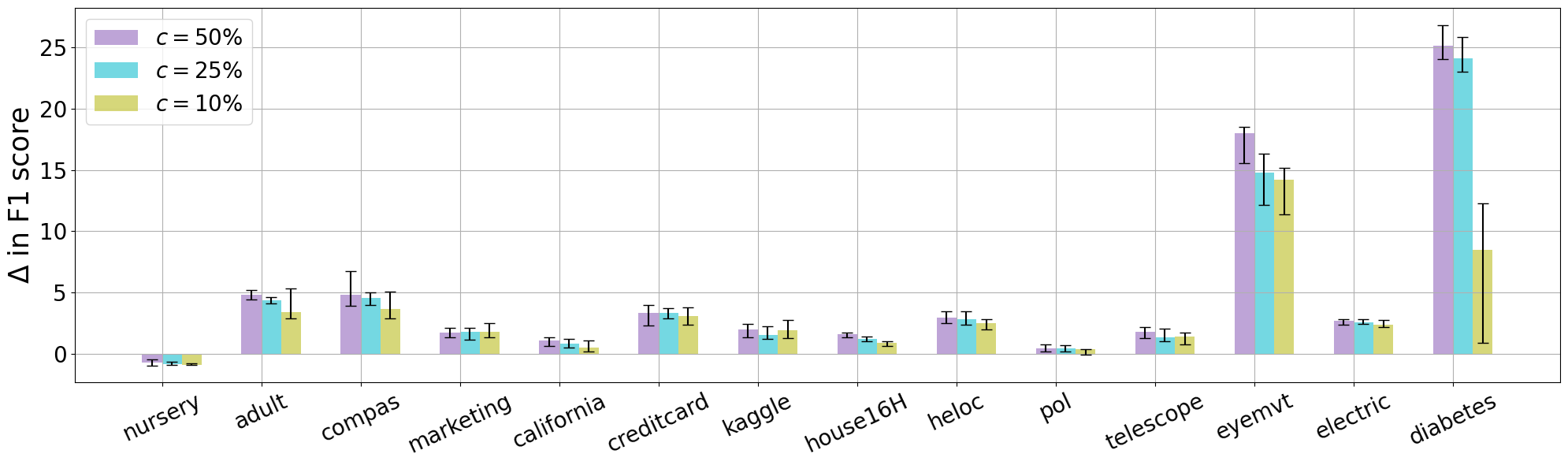}}
\hfil 
\subfloat[IBM trees]{\includegraphics[width=\linewidth]{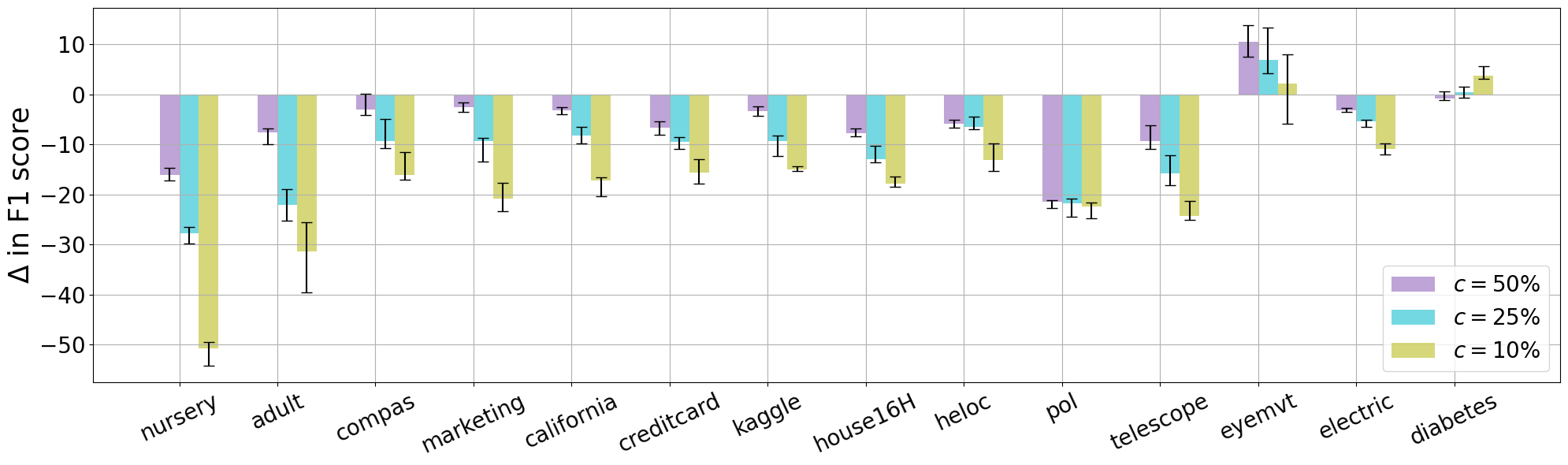}}
\caption{\footnotesize{$\Delta$ in $F1$ score across random train/test splits in the case of partial client participation for FedGA and IBM trees.}}
\label{fig:classification_iid_partial}
\end{figure}

In addition to the total number of clients, we investigate the impact that varying the percentage of client participation has on training outcomes. Fig. \ref{fig:classification_iid_partial} highlights the robustness of our GA-based method and the drawback of DP-based IBM trees as we vary the percentage of active clients per generation $c \in \{50\%, 25\%, 10\%\}$. Note that the latter leverages encryption and SMC to reduce the amount of noise required for $\epsilon$ privacy, thus improving accuracy. As a result, the reduction is proportional to the number of active clients in the training procedure. As fewer clients contribute, we observe a significant decrease in predictive power. Meanwhile, FedGA exhibits much more stability for various levels of client participation.

\subsubsection{Regression}
\begin{figure}[!t]
\centering
\subfloat[$\Delta$ in $MSE$ of federated trees compared to local trees.]{\includegraphics[width=\linewidth]{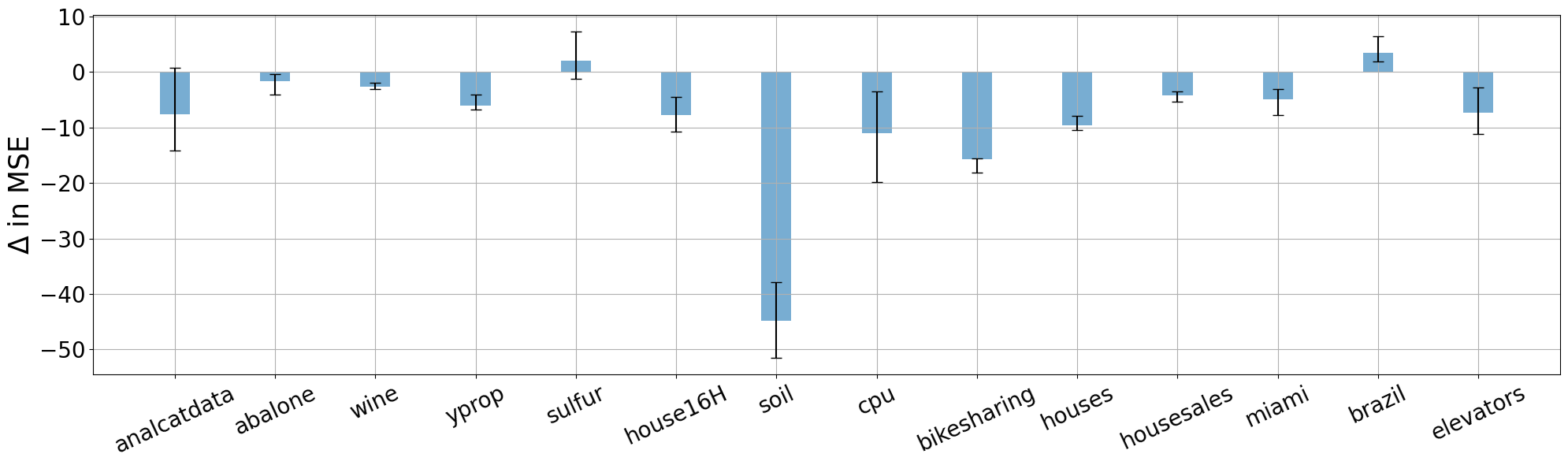}}
\hfil
\subfloat[Percentage of clients for which the federated trees outperform local trees.]{\includegraphics[width=\linewidth]{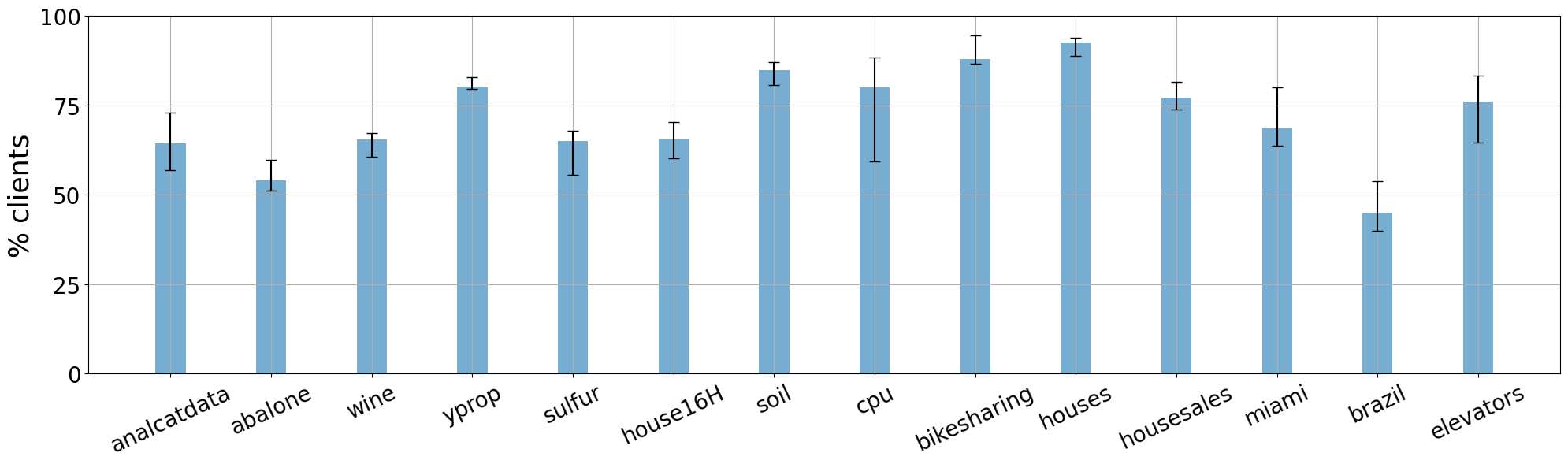}}
\caption{\footnotesize{Performance of federated trees compared to local trees for different datasets across 10 random train/test splits. (a) The medians and IQRs of $\Delta$ in $MSE$. (b) The median percentages of clients for which federated trees perform better as well as the corresponding IQRs. The $\Delta$'s and percentages of clients are averaged across independent runs for each train/test split.}}
\label{fig:regression_iid_full}
\end{figure}
Finally, we report the results for the regression data sets, and since the test score is $MSE$, a negative value for $\Delta$ is preferred. Similarly to classification, we observe in Fig. \ref{fig:regression_iid_full} that FedGA trees consistently outperform local trees for most datasets ($10$ out of $14$) and for at least $50\%$ of the clients. For the two datasets, \texttt{sulfur} and \texttt{brazil}, although we have more than half of the clients making improvements with federated trees, it is likely that $\Delta$ is skewed by the increase in $MSE$ for other clients. 

\begin{figure}[!t]
\centering
\subfloat[Number of decision nodes.]{\includegraphics[width=\linewidth]{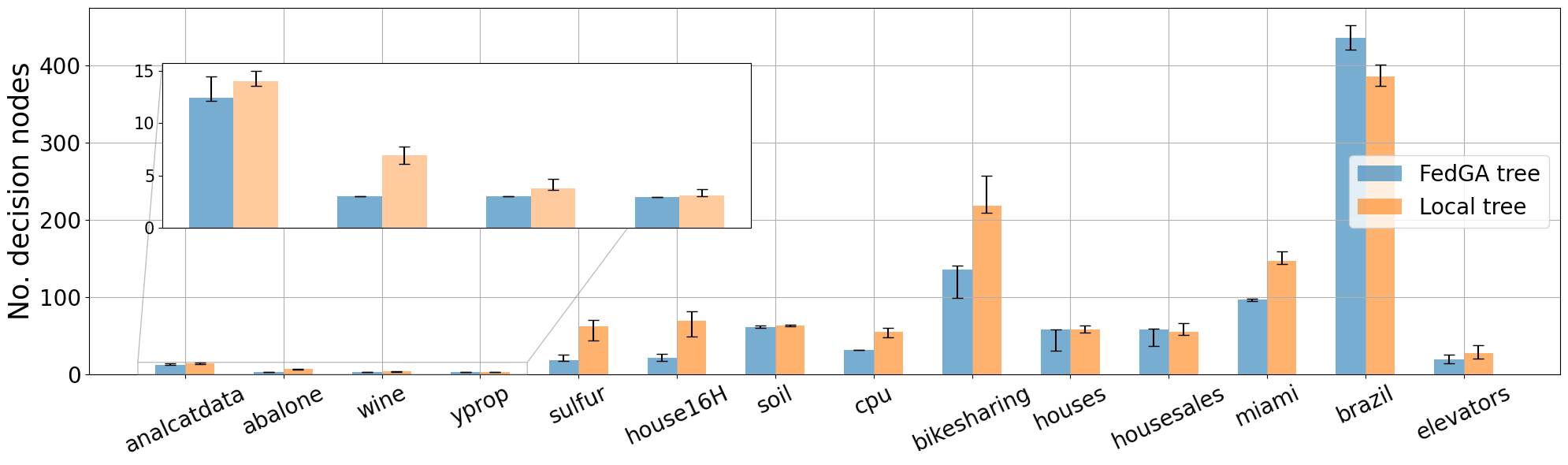}}
\hfil
\subfloat[Normalized tree edit distance between clients’ FedGA trees and their
corresponding local trees.]{\includegraphics[width=\linewidth]{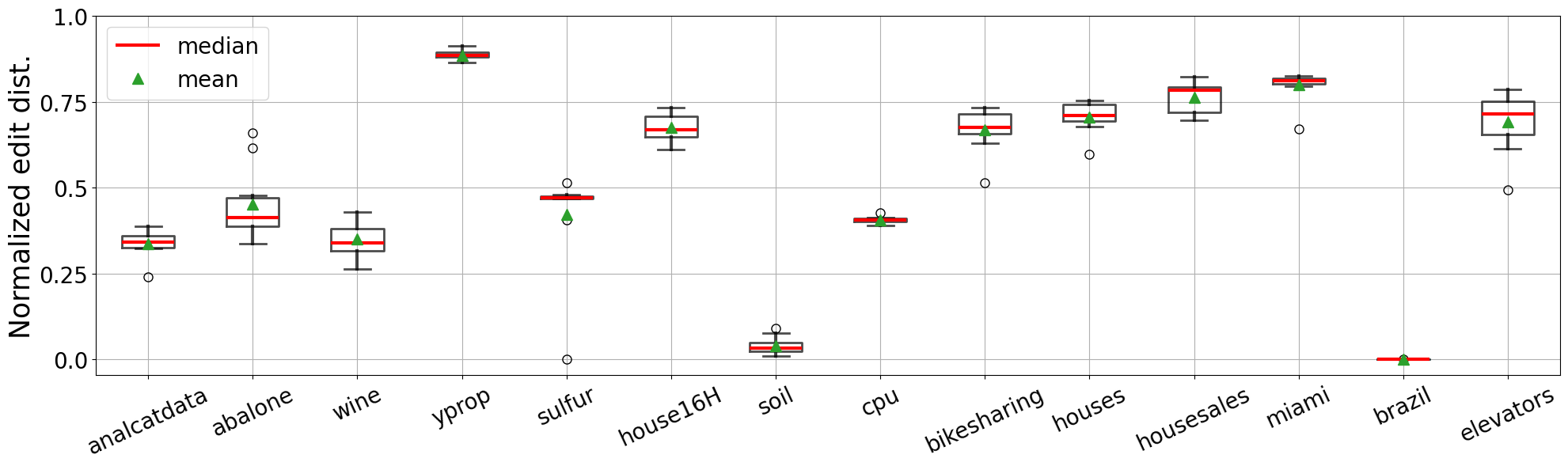}}
\caption{\footnotesize{Comparison of tree structures for different datasets across 10 random train/test splits. (a) The medians and IQRs of number of decision nodes averaged across clients and independent runs of FedGA. (b) The normalized tree edit distance between clients’ FedGA trees and their corresponding local trees, averaged across independent runs.}}
\label{fig:regression_iid_full_tree}
\end{figure}

We also inspect the complexity of FedGA trees in Fig. \ref{fig:regression_iid_full_tree} and find that the average number of decision nodes in FedGA trees is lower than that of local trees. On the other hand, we notice in Fig. \ref{fig:regression_iid_full_tree}(b) that there is an increase in similarity between FedGA and the corresponding GA local structures compared to classification. For the \texttt{brazil} dataset, almost all clients end up with the local structures of the initial population, yet we observe an increase in $MSE$. This is likely because the benchmarks are trained with locally optimized maximum depths while the local structures in the initial GA population are obtained with the aggregated maximum depth determined by the server. It should also be noted that the \texttt{brazil} dataset is the only one for which the elitism strategy is activated for every train/test split.

\begin{figure}[!t]
\centering
\subfloat[Varied total number of clients in the system.]{\includegraphics[width=\linewidth]{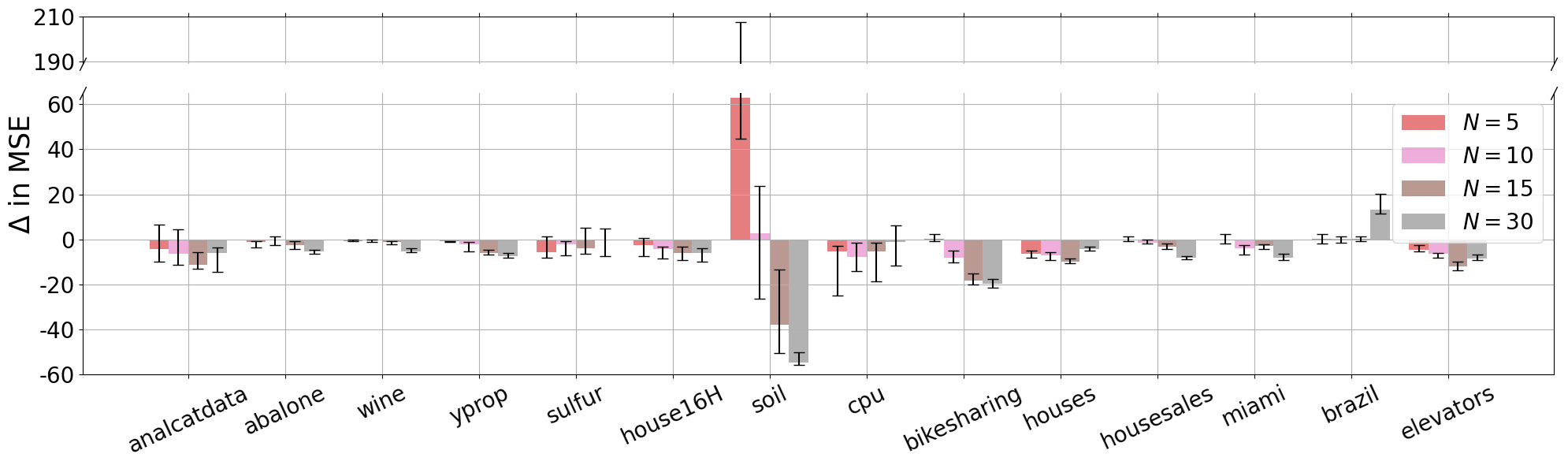}}
\hfil
\subfloat[Varied percentage of active clients.]{\includegraphics[width=\linewidth]{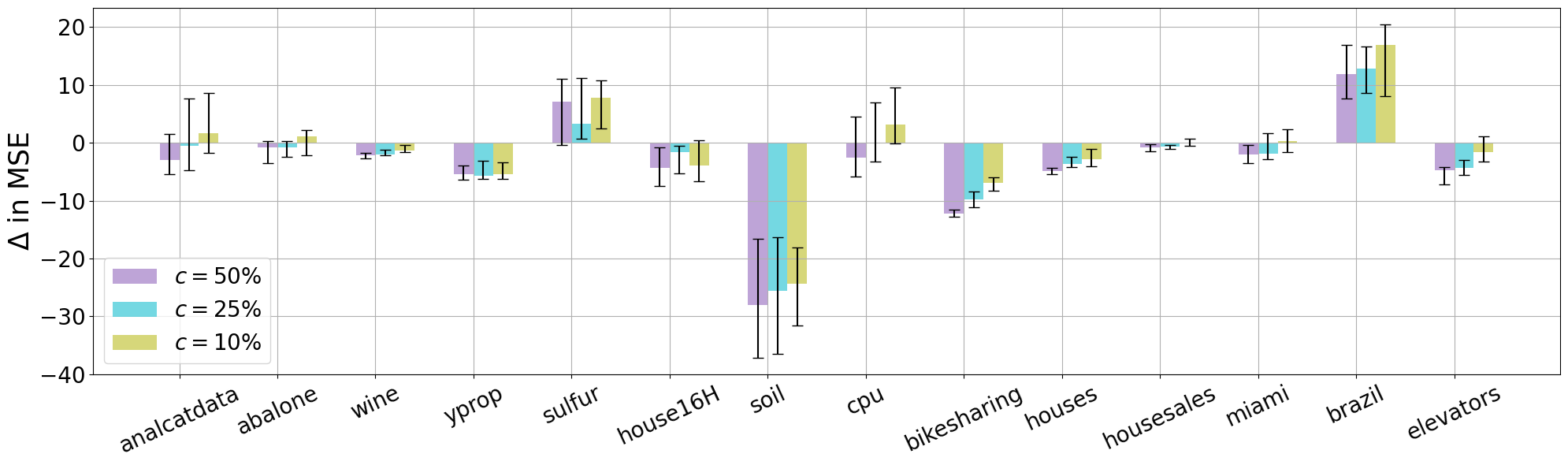}}
\caption{\footnotesize{$\Delta$ in $MSE$ across random train/test splits for FedGA as the total number of clients and the percentage of active clients varies.}}
\label{fig:regression_iid_partial}
\end{figure}

Last but not least, we investigate the performance of FedGA when the total number of clients and the percentage of active clients vary. We report the results in Fig. \ref{fig:regression_iid_partial}. Compared to classification, we observe that FedGA trees generally perform worse than local trees when there are fewer clients in the FL system, particularly for the \texttt{soil} dataset when $N=5$. Upon closer inspection, we find that there are still on average $50\%$ of the clients for whom FedGA trees perform better than local trees. However, for the remaining clients who do not achieve any improvement, the $MSE$ gaps between the federated and local models are significant enough to skew the $\Delta$. Meanwhile, in the case of partial client participation, FedGA still maintains stable predictive power as the percentage of active clients per generation decreases.

\section{Discussion}
In this paper, we present a federated method that draws upon Genetic Algorithm to build personalized decision trees and accommodates categorical and numerical data, thus enabling both classification and regression trees. We compare our approach with three benchmarks: 1) global trees trained with aggregated data without any privacy protocols, 2) local trees that clients train solely on local data, and for classification, 3) IBM trees by Truex et al. \cite{truex2019hybrid} that are trained using a greedy algorithm and differential privacy. Our method consistently outperforms the last two benchmarks and exhibits much more stability than IBM trees in the case of partial client participation, in which only a subset of clients contribute in a training iteration. With regression datasets, FedGA also achieves great results, but it is clear that using an unbounded metric like $MSE$ can skew evaluation results. For future work, we want to further improve the performance of FedGA for regression by experimenting with bounded metrics such as $R^2$ score or relative absolute error. We also plan to explore the effect of heterogeneous data and implement a more rigorous privacy assessment to quantitatively measure the impact of membership inference attacks on our method.
\bibliographystyle{plain}
\bibliography{reference}
\appendices
\section{Data preprocessing}\label{appendix:preprocessing}
For classification datasets with numerical features, we apply the \texttt{KBinsDiscretizer} to divide the range of the feature into $5$ bins and convert numerical values into ordinal values. To determine the bin edges, we employ the `\texttt{quantile}' strategy so that each bin contains approximately the same number of samples. Note that \texttt{KBinsDiscretizer} automatically adjusts the number of bins if there are not enough samples in some bins. 
Should there be only $1$ bin, we switch to the `\texttt{uniform}' strategy with $2$ bins so that the bin edges are of equal width. The \texttt{KBinsDiscretizer} function is fitted to the training set only to avoid information leakage, and thus the preprocessing procedure is repeated for each train/test split of the dataset.

\section{Tuning maximum depth for IBM trees}\label{appendix:ibmdepth}
\begin{figure}[!t]
    \centering
        \includegraphics[width=1\linewidth]{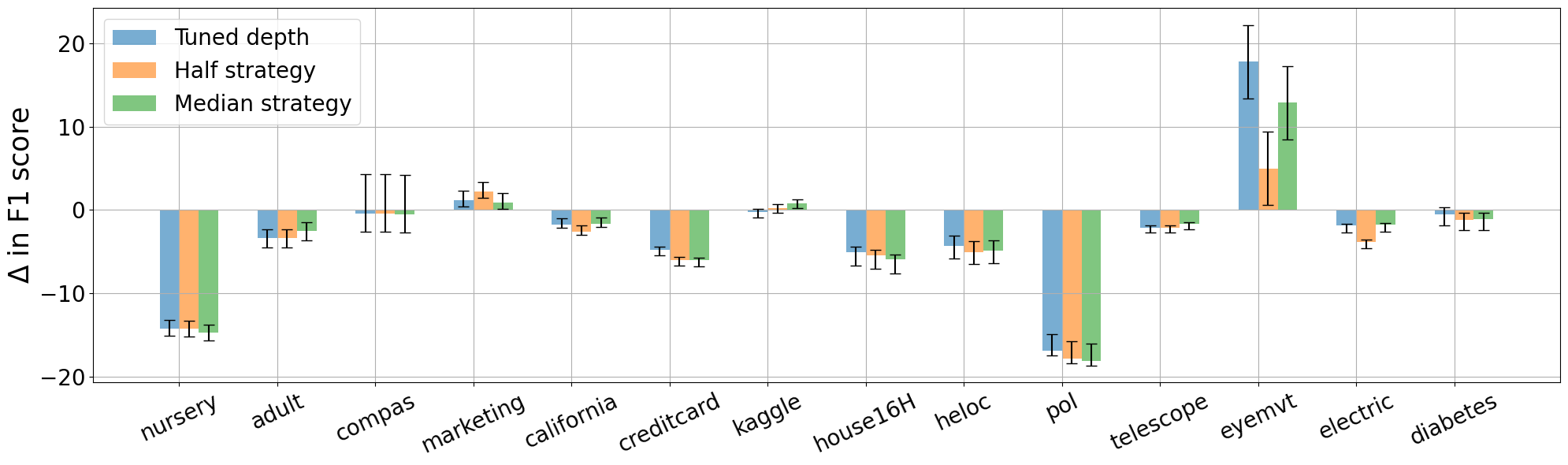}
        \caption{\footnotesize{$\Delta$ in $F1$ score across random train/test splits for IBM trees with different strategies for determining maximum depth.}}
        \label{appendix_fig:ibm_depth}
\end{figure}

We compare the performance of IBM trees with three different maximum depth strategies: fine-tuning, half strategy of \cite{truex2019hybrid}, and median strategy of FedGA. It is evident in Fig. \ref{appendix_fig:ibm_depth} that there is no major difference between the strategies, with the exception of the \texttt{eyemvt} dataset. However, tuned depth still has the best overall performance, with $\Delta$ averaged across all datasets being $-2.58$, while the average for the median strategy and half strategy is $-3.16$ and $-3.75$, respectively.

\end{document}